\documentclass[conference]{IEEEtran}
\IEEEoverridecommandlockouts
\usepackage{cite}
\usepackage{amsmath,amssymb,amsfonts}
\usepackage{algorithmic}
\usepackage{graphicx}
\usepackage{textcomp}
\usepackage{xcolor}
\usepackage{cancel}
\usepackage{url}
\def\BibTeX{{\rm B\kern-.05em{\sc i\kern-.025em b}\kern-.08em
    T\kern-.1667em\lower.7ex\hbox{E}\kern-.125emX}}

\newcommand{\RN}[1]{%
  \textup{\uppercase\expandafter{\romannumeral#1}}%
}
\begin{document}

\newcommand{\beginsupplement}{%
        \setcounter{section}{0}
        \renewcommand{\thesection}{S\arabic{section}}%
        \setcounter{table}{0}
        \renewcommand{\thetable}{S\arabic{table}}%
        \setcounter{figure}{0}
        \renewcommand{\thefigure}{S\arabic{figure}}%
        \setcounter{equation}{0}
        \renewcommand{\theequation}{S\arabic{equation}}%
     }

\title{Flow AM: Generating Point Cloud Global Explanations by Latent Alignment}

\author{\IEEEauthorblockN{1\textsuperscript{st} Hanxiao Tan}
\IEEEauthorblockA{\textit{AI Group} \\
\textit{TU Dortmund}\\
Dortmund, Germany \\
hanxiao.tan@tu-dortmund.de}
}

\maketitle

\begin{abstract}
Although point cloud models have gained significant improvements in prediction accuracy over recent years, their trustworthiness is still not sufficiently investigated. In terms of global explainability, Activation Maximization (AM) techniques in the image domain are not directly transplantable due to the special structure of the point cloud models. Existing studies exploit generative models to yield global explanations that can be perceived by humans. However, the opacity of the generative models themselves and the introduction of additional priors call into question the plausibility and fidelity of the explanations. In this work, we demonstrate that when the classifier predicts different types of instances, the intermediate layer activations are differently activated, known as activation flows. Based on this property, we propose an activation flow-based AM method that generates global explanations that can be perceived without incorporating any generative model. Furthermore, we reveal that AM based on generative models fails the sanity checks and thus lack of fidelity. Extensive experiments show that our approach dramatically enhances the perceptibility of explanations compared to other AM methods that are not based on generative models. Our code is available at: \url{https://github.com/Explain3D/FlowAM}
\end{abstract}

\section{Introduction}
Point cloud is one of the most important 3D computer vision data formats and is widely applied in navigation \cite{wang2021navigation,ilci2020high}, robotics \cite{yang2020novel,chen2020trajectory,yang2020teaser}, and medical domains \cite{yu20213d,cheng2020morphing}. With the proposal of PointNet \cite{qi2017pointnet}, the direct modeling of raw point clouds attracted numerous research interests, and to date refined point cloud neural networks have achieved satisfactory performances \cite{qi2017pointnet++,wang2019dynamic,ma2022rethinking}. However, due to the black-box property of neural networks \cite{castelvecchi2016can}, the reliability of these point cloud models cannot be warranted, especially in the aforementioned fields where human life is at stake. 

To alleviate the opacity of black-box models, explainability methods are proposed \cite{burkart2021survey}, which shed light on the rationale of model prediction by introducing additional modules or by simplifying the classifiers themselves. Explainability approaches can be classified into local and global approaches based on their field of view, where the former provide explanations for individual predictions \cite{adebayo2018local}, while the latter can generalize the decision-making patterns of an entire model or dataset \cite{saleem2022explaining}. Activation maximisation (AM) \cite{erhan2009visualizing,simonyan2013deep,nguyen2016synthesizing,zhou2017activation,mishra2019gan,borowski2020natural,fel2024unlocking,nguyen2017plug,bareeva2024manipulating} is one of the popular global explainability methods, whose idea is to maximise a specific neuron (typically the neuron in the last layer, representing a specific class) to visualise the desired input. One of the challenges of AM lies in the perceptibility of the explanations, e.g. in early attempts only nonsensical mosaics are generated \cite{erhan2009visualizing,simonyan2013deep}, which has been addressed in subsequent studies by incorporating regularization \cite{bareeva2024manipulating,fel2024unlocking} and generative models \cite{zhou2017activation,nguyen2017plug,mishra2019gan}.

Besides, the explainability of point cloud models has not been sufficiently studied. Existing research has mainly been conducted by transplanting explainability methods from images to point cloud models, including saliency maps \cite{gupta20203d,zheng2019pointcloud,figueiredo2020saliency}, surrogate models \cite{tan2022surrogate}, or critical point perturbations \cite{kim2021minimal,tan2023explainability,wicker2019robustness}. \cite{tan2023visualizing} is the only work that investigates AM approach for point clouds, which suggests that regularization-based AM fails to generate perceptible explanations, and therefore incorporates generative models to sample from real object geometries. However, we argue that the opacity of generative models themselves is one of the threats to the explanation transparency, and therefore the introduction of generative models is not an optimal alternative to the explainability methods. Additionally, generative models are inherently strong at reconstruction, which may raise concerns about information sources for generating explanations, thereby impairing the fidelity to the model to be explained.

In this work, we observe that different types of inputs activate various neurons within the model (activation flow) and propose a novel activation flow-based AM approach that regularizes the intermediate layers to yield explanations that approximate real objects. Our method significantly enhances the perceptibility compared to other non-generative model-based AM methods, and also demonstrates that even the simplest point cloud model learns the overall outline of objects through intermediate layers.

In summary, our contributions are as follows:
\begin{itemize}
    \item We propose a novel activation-flow based point cloud AM method to generate perceptible global explanations. To the best of our knowledge, this is the first method to generate perceptible global explanations for point clouds without incorporating generative models.

    \item We perform sanity checks on global explanations for point clouds, which expose that generative model-based AM approaches may suffer from concerns about the fidelity to the model to be explained.

    \item Extensive experiments qualitatively and quantitatively demonstrate that our approach dramatically enhances the perceptibility of explanations compared with other non-generative model-based AM approaches.
\end{itemize}
\section{Related Work}
In this section we introduce explainability methods, point cloud models, and explainability studies on point clouds.

\textbf{Explainability methods} typically refer to demonstrating the rationale for predictions made by a model through post-hoc processings. The most straightforward explanation are saliency maps, which are based on the gradient and highlight those regions (or features) of the input that play a key role in the prediction. The simplest saliency map calculates only the gradient of the input \cite{simonyan2013deep}, whereas subsequent studies have shown it to be biased, and corresponding refinements are proposed, such as Integrated Gradient \cite{sundararajan2017axiomatic} and SmoothGrad \cite{smilkov2017smoothgrad}. Nonetheless, saliency maps are not applicable for models where gradients cannot be computed. Another series of explainability methods consider the model as a black box and determine the importance of features by observing the correlation between the input and the output, known as surrogate model-based approaches \cite{ribeiro2016should,lundberg2017unified,ribeiro2018anchors,petsiuk2018rise}. All above methods generate explanations for a specific input and hence are also regarded as local explainability approaches. Instead, global explanations reflect to some extent the behavioral patterns of the whole model or dataset, typically represented by rule distillation \cite{pedapati2020learning,krishnan1999extracting} and Activation Maximization \cite{fel2024unlocking,bareeva2024manipulating}, etc.

\textbf{Activation Maximization (AM)} is a global explainability method that maximizes specific neurons by optimizing the inputs, and was first proposed by \cite{erhan2009visualizing}. Early AM-generated explanations contain massive high-frequency mosaics that are incomprehensible to humans. To improve the perceivability of the explanations, subsequent refinements employ L2-norm \cite{simonyan2013deep}, Gaussian Blur \cite{yosinski2015understanding} and Total Variation \cite{mahendran2016visualizing} to regularize the high-frequency pixels, or extract priors from the dataset to initialize the input vector \cite{mordvintsev2015inceptionism,nguyen2016multifaceted,wei2015understanding}. Incorporating generative models ensures sampling in the distributions of real objects, which significantly boosts the perceptibility of AM \cite{katzmann2021explaining,mishra2019gan,nguyen2017plug,nguyen2016synthesizing,zhou2017activation}. However, the opacity of the generative models themselves may instead render the explanations less plausible, and it is challenging to distinguish whether the explanations derive the information from the classification model or the generative one \cite{fel2024unlocking}.

\textbf{Point cloud networks} extract spatial geometric features from unorganized point sets and make predictions. The earliest model processing raw point clouds was proposed by \cite{qi2017pointnet}, which employs point-wise convolution and pooling layers to learn local and global features, respectively \cite{qi2017pointnet}. Subsequent studies incorporate local feature correlations \cite{qi2017pointnet++,liu2019relation}, tree structures \cite{riegler2017octnet,zeng20183dcontextnet}, and graphs \cite{simonovsky2017dynamic,wang2019dynamic} to boost the model performance, while deteriorating the opacity of the models.

\textbf{Point cloud explainability} is so far not adequately investigated. \cite{zheng2019pointcloud} identify points that are critical for prediction through perturbation, and other studies \cite{gupta20203d,tan2022surrogate} attempt to transplant explainability methods from images so that they are also applicable to point clouds. \cite{tan2023visualizing} is the pioneer in the study of point cloud AM, which demonstrates that traditional regularization (such as L2-norm) is incapable of generating perceptible explanations, and proposes an autoencoder-based approach that samples and optimizes explanations from real object distributions. Although generative models significantly enhance the perceptibility of explanations, we argue that their inherent opacity is not conducive to explainability, while it is difficult to determine from which model the generated profile information originated. We show salinity checks in our experiments to expose the fidelity concerns of AM based on generative models.

\section{Methods}
In this section, we introduce existing AM implementations, point out their flaws (Sec. \ref{intro_AM}), demonstrate the activation flow for different types of inputs (Sec. \ref{Activation_flow}), and elaborate on the proposed approach (Sec. \ref{flow_AM}).

\subsection{Activation Maximization (AM) and regularization} \label{intro_AM}
\begin{figure*}
    \begin{centering}
    \includegraphics[width=0.7\textwidth]{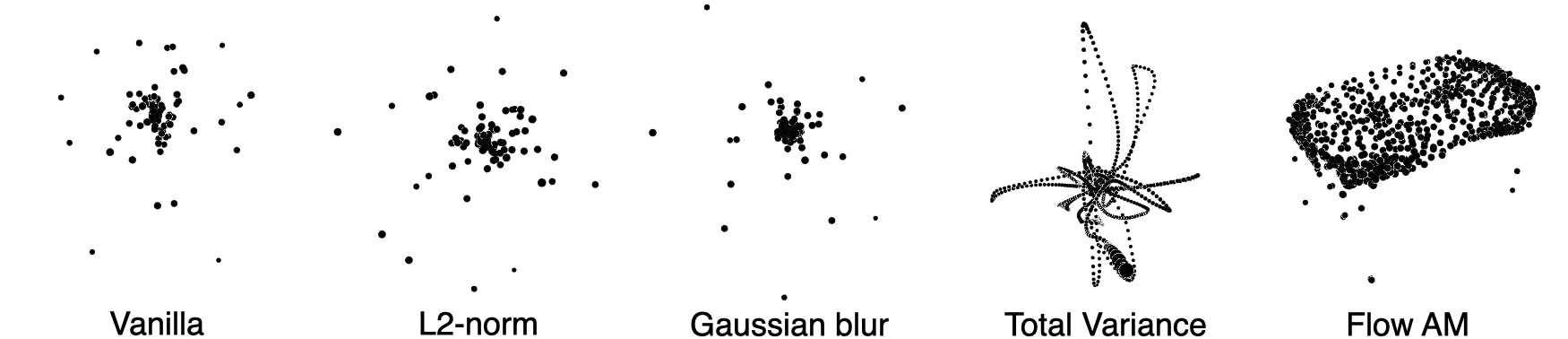}
    \caption{Visualization comparison of non-generative model-based point cloud AM approaches with the target label ``Table". From left to right : Vanilla AM, regularized by L2-norm, with Gaussian blur applied, regularized by Total variance and our flow-based AM.}
    \label{Fig:AM_compare}
    \end{centering}
\end{figure*}

AM is a neural network visualization technique that demonstrates which inputs maximally activate target neurons. AM can be formulated as
\begin{equation} \label{eq:am}
x^* = \underset{x}{\mathrm{argmax}}\, (a_i^l(F,x))
\end{equation}
where $x$ denotes the input, $F$ denotes the model to be explained, and $\alpha^l_i$ denotes the $i^{th}$ neuron on the $l^{th}$ layer of the model. Typically, $l$ is chosen as the logits or softmax layer so that $a_i^l$ represents a specific category. \cite{erhan2009visualizing} utilizes gradient ascent to optimize the input $x$. However, in initial attempts, the gradient is more sensitive to those high-frequency pixels, and thus the generated explanations are mostly mosaics that are unrecognizable to humans. To address this property, subsequent studies propose various solutions to regularize high-frequency pixels, such as L2-norm \cite{simonyan2013deep}, Total Variation \cite{mahendran2016visualizing} and Fourier transform \cite{fel2024unlocking} (See Sec. \ref{sec:tradictionAMformular} for regularization formulas). Nevertheless, in point cloud AM, the above solutions fail to achieve improved perceivability due to the structures of the point cloud models, which are specifically designed for disordered point sets, do not contain multi-size convolution kernels, and thus regularizing neighboring pixels is no longer applicable \cite{tan2023visualizing}, as illustrated in Fig. \ref{Fig:AM_compare}. In addition, existing studies reveal that gradient ascent algorithms for point clouds merely cause individual (critical) points to expand outward \cite{tan2023visualizing,tan2023explainability}, which is distinct from high-frequency pixels in images. 

\subsection{Activation flow} \label{Activation_flow}
All existing AM methods so far focus only on individual target activations, however, we observe that for point cloud models, neurons in the intermediate layer are instead crucial for learning the overall outline of objects.

\begin{figure*}
    \begin{centering}
    \includegraphics[width=1.0\textwidth]{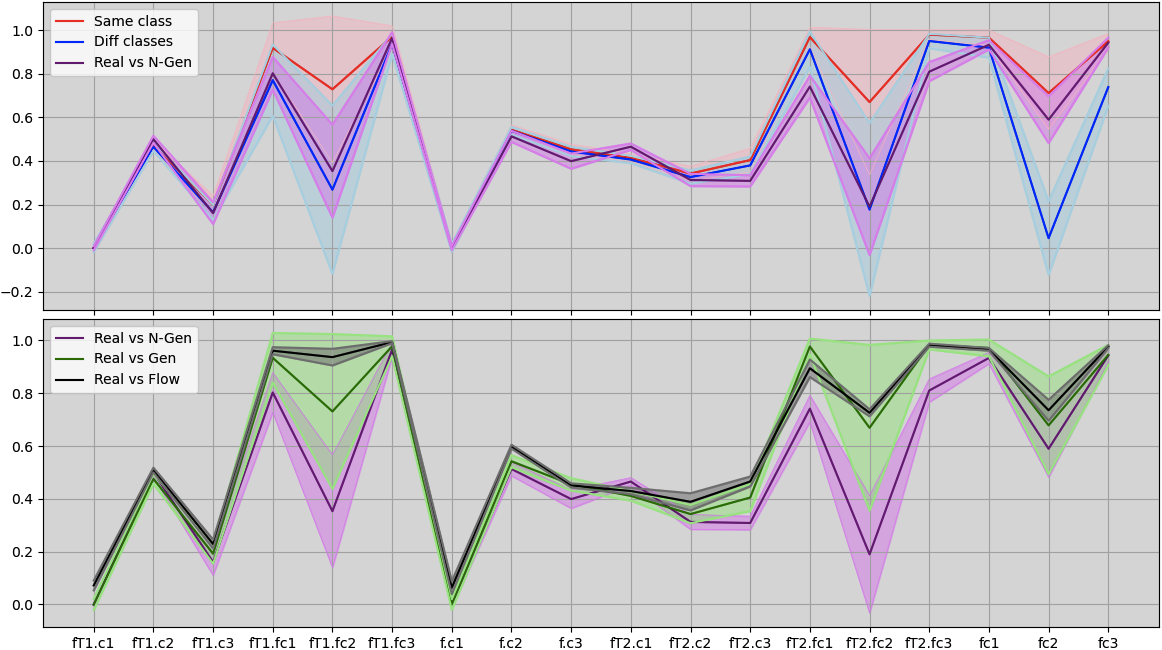}
    \caption{Upper part: similarities in the activation of neurons within the model when predicting real objects of the same (red) and different classes (blue), respectively. The purple curve indicates the activation similarity between non-generative AMs and real objects. Lower part: similarities between the activation of neurons within the model when predicting real objects and when predicting explanations generated by non-generative model-based (purple), generative model-based AM (green) and our flow-based AM (black), respectively. The y-axis indicates the cosine similarity and the x-axis is the name of each layer of PointNet. We utilize cosine similarity to calculate the activation likelihood of each layer, which yields similar results with other metrics such as Pearson and Spearman's coefficients.}
    \label{Fig:flow_compare}
    \end{centering}
\end{figure*}

To elaborate, we start with a simple experiment demonstrating the activation flow in the model. We chose PointNet as the predictor and ModelNet40 as the experimental dataset. We select arbitrary two inputs belonging to the same class from each of the 40 categories and predict them with the model. We record the activation of neurons at each layer in the model at the time of prediction and compute the similarity of neuron activation when processing inputs of the same class. We simultaneously calculate the similarities of the activations when processing inputs from different classes. The comparison is shown in the upper part of Fig. \ref{Fig:flow_compare}. It can be observed that, except for the logits layer, there are significant discrepancies between neuron activations in parts of the intermediate layers (mainly the fully-connected layers, e.g. $fT1.fc1$ and $fT2.fc2$) when processing inputs from the same and different classes. In the lower part of Fig. \ref{Fig:flow_compare}, we exhibit the similarity of neuron activations when the model predicts explanations generated by AMs based on non-generative and generative models. Despite the high similarity of the logits layers, there are remarkable dissimilarities in the activation of the intermediate layers, which we believe is the pivotal cause of the inability of non-generative AM to yield perceptible explanations. Another evidence is that when observing the similarity between non-generative AM and real objects (purple and blue curves in the upper part), we find it is remarkably close to the activation similarity at the intermediate layers while predicting different classes of objects (except for the last two layers). This suggests that two objects with completely different profiles may obtain similar predictions by gradient ascent, which explains why attacking algorithms can always succeed in deceiving the model by perturbing only few number of points \cite{kim2021minimal,tan2023explainability,zheng2019pointcloud}. Overall, traditional non-generative model-based AM fails to take into account the overall outlines of the inputs, and thus is unable to generate human-perceivable explanations.

\subsection{Flow AM} \label{flow_AM}
\begin{figure*}
    \begin{centering}
    \includegraphics[width=0.8\textwidth]{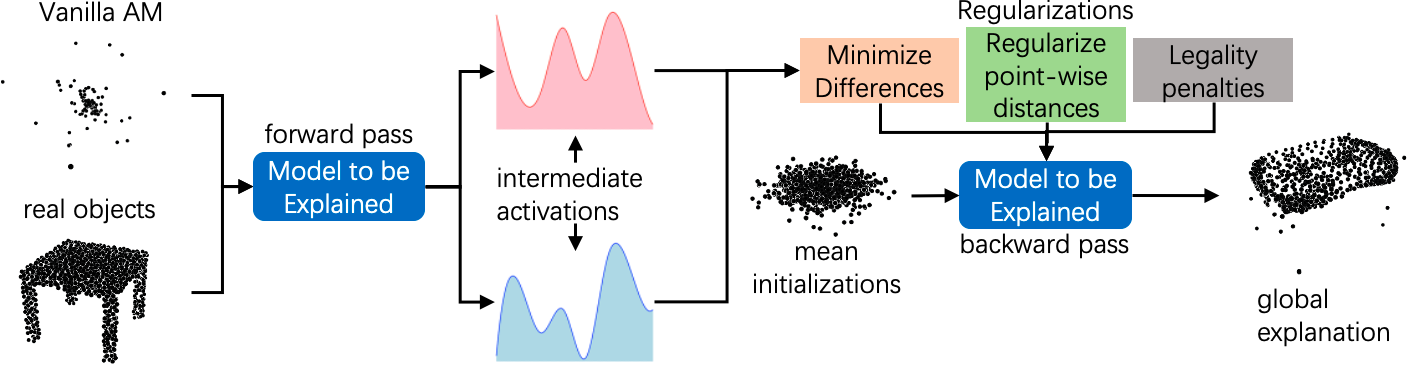}
    \caption{Overview of the structure of Flow AM. Raw AM represents the explanations generated by AM without any regularization.}
    \label{Fig:Flow_overview}
    \end{centering}
\end{figure*}

To address the aforementioned issues, we propose a novel method for generating AM explanations without generative models, which we call Flow AM. Flow AM is based on the simple idea that when generating explanations, we force those intermediate layers responsible for learning the global profiles to perform analogously when predicting real objects of the same class during gradient ascent. The general process of Flow AM is: A) Generate AM explanations without any regularization. B) Predict the raw AM and real objects respectively with the model and record the activation distributions of the intermediate layers. C) Filter the intermediate layers with large discrepancies to be the target of regularization, and define a synthesized loss. D) Add other geometric regularizations and fine-tuning hyperparameters. We illustrate an overview of the structure of Flow AM in Fig. \ref{Fig:Flow_overview}.

The regularizations we incorporate for AM are grouped into the following four types.

\textbf{\RN{1}. Activating class neuron.} Activating neurons of the target category (AM loss) is the most basic AM loss term, which forces the explanations to highly activate a particular neuron on the logits or Softmax layer to achieve representativeness. AM loss can be formulated as
\begin{equation}
    L^c_{AM} = -\alpha_c^{out}
\end{equation}
where $\alpha_c^{out}$ denotes the $c^{th}$ neuron (the $c^{th}$ class) on the output layer.

\textbf{\RN{2}. Latent approximation of real objects.} We show in Sec. \ref{Activation_flow} that a portion of the neuron activations in the intermediate layers are determined by the outline of the object. Therefore, during the gradient ascent, the neuron activations in the intermediate layers of the explanations should be aligned with real objects as much as possible.

\begin{itemize}
    \item[] \textbf{\RN{2}.A Structural analysis of point cloud models.} To minimize the variation in the intermediate layers, the layers that learn the contours are required to be chosen from the model. The vast majority of point cloud models share the common general architecture, consisting of local point features, pooling layers, and global latent features. For simplicity, we start by analyzing the structure of PointNet \cite{qi2017pointnet}. Being one of the simplest point cloud networks in terms of structure, the layers of PointNet are grouped into two primary categories, i.e., convolutional and fully connected layers. They are uniformly distributed in the three modules of the model which are T-net1 (\textit{fT1} in the x-axis of Fig. \ref{Fig:flow_compare}), T-net2 (\textit{fT2}) and the main network (\textit{f.c} and \textit{fc}). The convolutional layers can be excluded from alignment, as all convolutional kernels in the network are $1\times 1$, which learns only point-wise features. As can be observed in Figure 2, the activation of neurons on the convolutional layers (ending in ``$.ci$" where $i=1,2$ and $3$) varies widely even when predicting the identical class of inputs (average cosine similarity$< 0.6$). Instead, structurally, each fully-connected layer contains global information (including fT1 and fT2, as fT1.c3 and fT2.c3 are followed by a global pooling layer), and thus the these layers are considered to be the primary targets for alignment. Furthermore, the activations of neurons on (part of) the fully connected layers yield statistically noticeable distinctions when predicting inputs from same and different classes, which serves as evidence that they contain contour information related to the class.

    \item[] \textbf{\RN{2}.B Latent alignment for non-generative AM.} Under the above prior, we observe that when predicting the explanations of non-generative AM (purple and red curves in the upper and lower part of Fig. 2, respectively), the activation differences with respect to real objects are mainly in $fT1.fc1$, $fT1.fc2$, $fT2.fc1$, $fT2.fc2$ and $fc2$. Therefore, we incorporate these differences into the regularization of Flow AM, forcing the neurons in the intermediate layers to align their activation distributions with the real objects. 
    
    We first predicted each object in the testset with the model and record the activations of the aforementioned intermediate layers. Subsequently, we aggregate these records according to category and calculate the average activation for each class, denoted as $\omega^l_{c}$, where $l$ denotes the $l^{th}$ layer and $c$ denotes the class. The latent alignment regularization in Flow AM is formulated as

    \begin{equation}
        L^c_{LA}=-\sum_{l \in l_f}\alpha_l \times SC(\omega^l_{c}, \tilde{\omega}^l)
    \end{equation} \label{eq:latent_approx}
    where $l_f$ denotes set of layers to be aligned, $\alpha_l$ denotes the regularization weight on layer $l$, $SC$ denotes the cosine similarity and $\tilde{\omega}^l_{c}$ denotes the neuron activation on layer $l$ during the forward propagation of the optimization process towards class $c$. So far, we have not found an optimal strategy for determining $\alpha_l$ for various models, hence the setting of $\alpha_l$ is heuristic.
\end{itemize}

By regularizing the activation flow, we enable the activations in the intermediate layers for non-generative model-based explanations to be analogous to real objects (See the black curve in the lower part of Fig. \ref{Fig:flow_compare}). 

\begin{figure}
    \begin{centering}
    \includegraphics[width=0.25\textwidth]{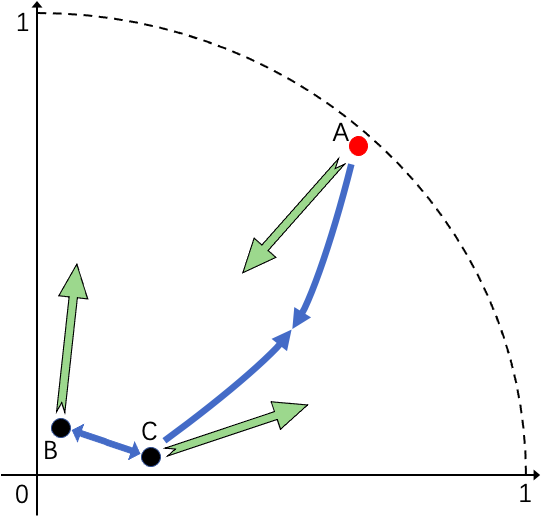}
    \caption{The intuition of the point continuity loss. Point A is a critical point that tends to expand during AM, while points B and C are common points that are ignored by the gradient. The outer dashed curve is the boundary of a legal point cloud instance. The green and blue arrows indicate the direction of two different terms in the continuity loss, i.e., the distance to the origin and to the nearest neighboring point, respectively.}
    \label{Fig:c_s_loss}
    \end{centering}
\end{figure}

\textbf{\RN{3}. Point continuity and smoothness.}
The point cloud AM tends to expand individual points outward during the optimization process, which causes the explanation to suffer from a small number of outliers that are severely off-center (which are the critical points that are decisive for the prediction \cite{tan2023explainability}), with most of the rest (common points) clustered near the origin \cite{tan2023explainability,tan2023visualizing} (As shown by Vanilla, L2-norm and Gaussian blur in Figure \ref{Fig:AM_compare}). We mitigate the issue in two aspects: limiting the expansion of outlying critical points and diminishing their distance from neighboring points, while shifting the rest of points away from the origin and spreading them out by increasing their distance from neighboring ones. Our continuity loss can be formulated as

\begin{equation}
L_{C}=\sum_{i=1}^{n_p} \left |B-(min\{\left \| p_i-p_j \right \|_2\}_{j\neq i} + \left \| p_i \right \|_2)  \right |
\end{equation} \label{eq:continuity}
where $n_p$ and $p_i$ denote the total number of points and the $i^{th}$ point in the explanation, $min\{\left \| p_i-p_j \right \|_2\}_{j\neq i}$ is the distance from $p_i$ to its nearest neighboring point, and $B$ is the legal boundary of the dataset. The intuition behind continuous loss is illustrated in Fig. \ref{Fig:c_s_loss}. Each point is impacted by two regularizations in the optimization process, i.e., absolute and relative distances, the former being $\left \| p_i \right \|_2$ in Eq. \ref{eq:continuity} which is the distance to the origin, and the latter being $min\{\left \| p_i-p_j \right \|_2\}_{j\neq i}$. For example, the location of the critical point A in the figure is close to the legal limit ($B=1$) of point cloud instances, with an absolute distance close to 1, as well as a large relative distance since the critical point is sparse. The optimizer reduces both its absolute and relative distances to relocate it back to the origin (the green arrow) and closer to its neighboring point C (the blue arrow). In contrast, the absolute and relative distances of the common point C are relatively small, and the optimizer raises them up so that the sum approaches 1. By balancing the two distances, points are evenly distributed on the surface of the generated explanations to enhance their perceptibility.

\textbf{\RN{4}. Legal restriction.} Point cloud datasets are subject to a legal spatial limitation, e.g., ModelNet40 is legal in the interval [-1,1], and instances containing points that exceed this limitation are treated as illegal inputs. To guarantee that most of the explanations are valid, we incorporate a simple regularization that restricts part of the critical points from exceeding the legal limitation during the AM process, which is frequently observed in point cloud AMs without regularization \cite{tan2023explainability}. The legal restriction is formulated as
\begin{equation}
    L_{LR}=\sum_{i=0}^{n_p} ReLU(\left | p_i \right |-C)
\end{equation}
where $C$ denotes the legal constrain of the dataset.

Finally, we aggregate the aforementioned loss terms as the optimization target for Flow AM. In practice, we observe that $L_{C}$ and $L^c_{LA}$ are conflicting to some extent, rendering the optimization non-converging. We incorporate a trade-off coefficient $\beta$ for these two terms, which is adjusted to balance the smoothness and closeness to the real object of the explanations. The final loss function of Flow AM can be formulated as
\begin{equation}
    L^c_{Flow} = L^c_{AM} + (1-\beta)L^c_{LA} + \beta L_{C} + L_{LR}
\end{equation}

In addition to the loss terms, we employ the average initialization to improve the quality of the explanation \cite{mordvintsev2015inceptionism}. We compute the average of the points in the test set for all objects by category and utilize it as the input to Flow AM, which improves the efficiency of the optimization process and the quality of the explanations. 

\section{Experiments}
In the experiments, we select ModelNet40 and PointNet as the main experimental datasets and models, respectively. ModelNet40 is one of the most popular datasets, which consists of 40 classes totaling 12,311 CAD models, where the training and test sets contain 9,843 and 2,468 object instances, respectively. PointNet is the pioneer of raw point cloud models, which is mainly composed of point-wise convolutional, pooling and fully connected layers, with a relatively straightforward structure, and is ideal for explainable analyses. We set the regularization weights in Eq. \ref{eq:latent_approx} to be $N_l\times1e-9$, $1e-9$, $1e-1$, $1e-1$ and $1$ for $fT1.fc1$, $fT1.fc2$, $fT2.fc1$, $fT2.fc2$ and $fc2$, respectively, where $N_l$ is the total number of activations contained in this layer. In addition, we set $\beta=0.1$ to balance the latent and point-wise distances. As a reference, we choose the regularization methods in Fig. \ref{Fig:AM_compare} as well as the generative model-based approaches proposed by \cite{tan2023visualizing} as competitors, which to the best of our knowledge are currently all the approaches that have been attempted by existing researches. The number of optimization iterations for each method is set such that the target activation almost converges.

\subsection{Qualitative comparison of global explanations}

\begin{figure*}
    \begin{centering}
    \includegraphics[width=0.8\textwidth]{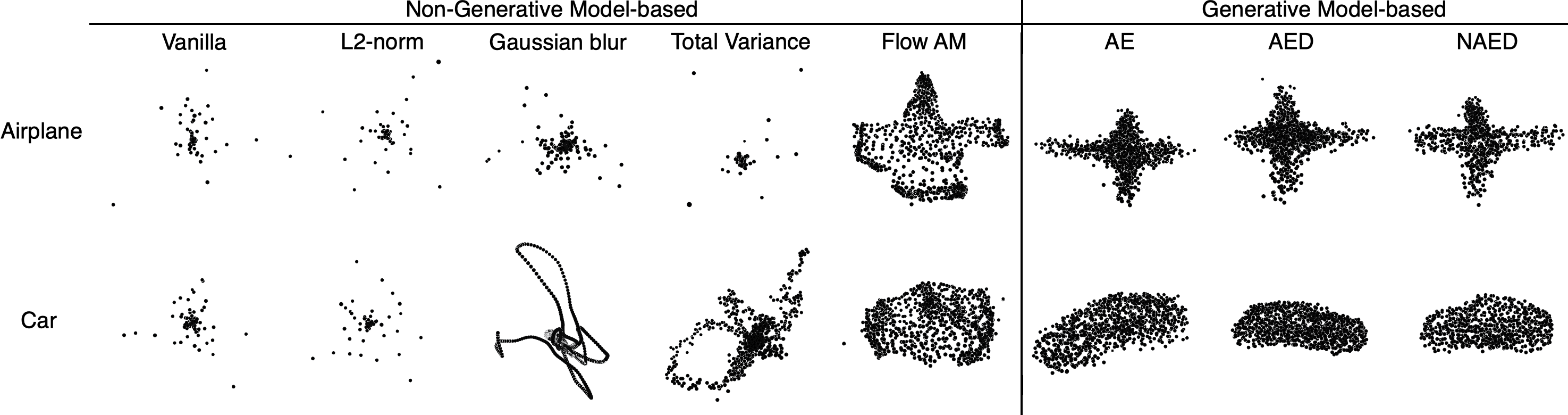}
    \caption{Qualitative comparison of non-generative model-based global explanations. From left to right are, no regularization, L2, Gaussian blur, total variance, flow regularizations (ours), AE \cite{tan2023visualizing}, AED \cite{tan2023visualizing} and AED \cite{tan2023visualizing}. Note that the first five approaches are based on non-generative models while the last three are based on generative models. More visualizations can be seen in Fig. \ref{Fig:more_visu_compare}.}
    \label{Fig:Qualitative_Flow_compare}
    \end{centering}
\end{figure*}

We select several common classes as target activations from ModelNet40 and visualize them utilizing the aforementioned AM methods. Figure \ref{Fig:Qualitative_Flow_compare} illustrates the performance of existing regularization approaches and generative model-based AM methods. It can be observed that among the non-generative model-based AM methods, none of them, except Flow AM, are able to generate global explanations that can be perceived by humans. The vast majority of regularizations are aimed at eliminating high-frequency points, while the issue with vanilla point cloud AM is the high sensitivity of individual points to the target activation. With the incorporation of latent flows in the intermediate layers, our Flow AM successfully reconstructs the global profiles of objects and achieves almost comparable quality to that based on generative models with the help of continuity regularization. Furthermore, though generative model-based approaches yield high-quality explanations as well, our Flow AM does not rely on any external priors, guaranteeing the fidelity of explanations to the model to be explained, while dramatically diminishing the computational intensity.

\begin{figure}
    \begin{centering}
    \includegraphics[width=0.475\textwidth]{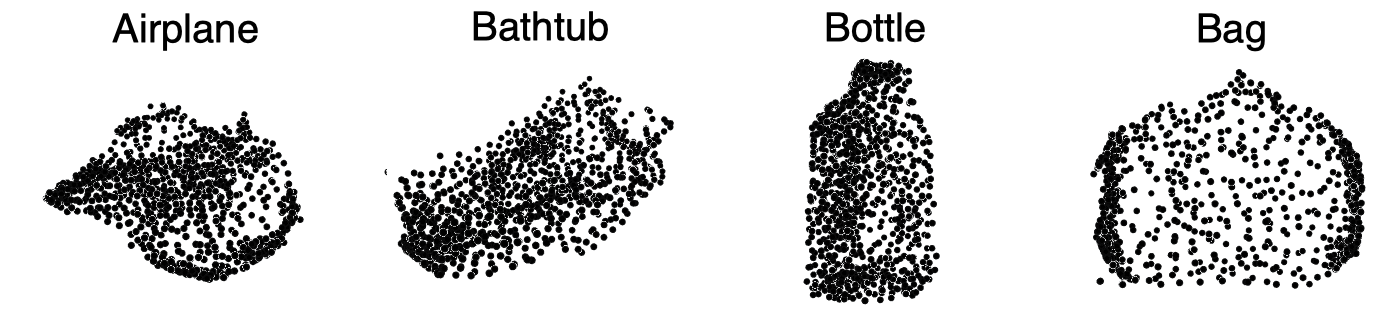}
    \caption{Global explanations for PointNet trained on ShapeNet dataset.}
    \label{Fig:Qual_shapenet}
    \end{centering}
\end{figure}

We also generate global explanations for PointNet trained on ShapeNet, and select the first few categories to display in Fig \ref{Fig:Qual_shapenet}. Note that the wing contours of the airplane are not as clear as those of ModelNet40, this is because the airplane class in ShapNet dataset contains a large percentage of delta-winged airplanes, which is the feature learned by the latent activation of the model.

\subsection{Quantitative evaluations}
For an objective comparison of AM methods, we quantitatively evaluate the quality of the generated global explanations in the following three metrics.

\textbf{Representativity} reflects the ability to generalize about a particular category, i.e., the more distinctive the characteristics of the category that the explanation possesses, the more representative it is. Note that the representativity evaluator is the model to be explained, rather than humans. There are features that are semantically unavailable to humans but can highly activate higher-level activations of the model, such as the high-frequency mosaics in early feature visualizations \cite{simonyan2013deep}. We simply measure the neurons corresponding to specific classes in the Logits and Softmax layers as the metric for representativity assessment, where the former is considered to be absolute representativity, which is the absolute magnitude of activation of the neuron, while the latter is considered to be relative representativity, which takes the activation magnitudes of all the classes into account and calculates the class probability.

\textbf{Point-wise perceptibility} denotes the discrepancy between the explanations and real objects from the same class in the three-dimensional geometric space. Intuitively, the more proximate an explanation is to the contours of real objects, the easier it is for humans to recognize it, and therefore the more explainable it is. We employ Chamfer Distance (CD) as point-wise perceptibility metric, which can be respectively formulated as:

\begin{equation}
    CD(x_g,x_r)=\frac{1}{|x_g|}\sum_{p_m\in x_g}\min_{p_n\in x_r}\left \| p_m-p_n \right \|_2
\end{equation}
where $x_r$ and $x_g$ are real object samples and generated explanations, respectively.

\textbf{Latent perceptibly} is the discrepancy between explanations and real objects when encoded as features in the intermediate layers of the classifier. Existing research reports that the variations in specific intermediate layers strongly reflects the differences in human perceptions of input \cite{heusel2017gans}. We employ Fréchet inception distance (FID) as the metric of latent perceptibility and follow existing point cloud applications \cite{sun2020pointgrow,tan2023visualizing} in selecting global features of PointNet as the recording layer for latent distances. FID is formulated as

\begin{equation}
    FID(x_g,x_r)=\left \| \mu_r -\mu_g  \right \|^{2}+Tr(\sigma_r + \sigma_g - 2(\sigma_r \sigma_g)^{\frac{1}{2}})
\end{equation}
where the latent vectors $l_r\sim \mathcal{N}(\mu_r,\sigma_r)$ and $l_g\sim \mathcal{N}(\mu_g,\sigma_g)$ are approximated as normal distributions and $\mu$ and $\sigma$ are its mean and variance, respectively. When evaluating point-wise and latent perceptibilities, we randomly selecte five real objects from the same class in the dataset and calculate the corresponding distances between them and the generated explanations, respectively.

Quantitative evaluations are reported in Table \ref{tab:quantitative_eva}. It can be observed that methods based on generative models typically suffer from poor representativity, which is attributed to the additional prior embedded in the gradient by the generative model. In contrast, while methods based on non-generative models significantly enhance the magnitude of target activations, they perform poorly in terms of perceptibility (CD and FID), which is consistent with the conclusion shown in the qualitative comparisons. Our Flow AM strikes a better balance, preserving the representativeness of the non-generative approaches at the expense of an acceptable level of perceivability. 

Besides ModelNet40 , we also test the performance of Flow AM on ShapeNet and report the results in Table \ref{Tab:quantitative_shapenet}. Again choosing non-generative methods as competitors, Flow AM also dominates on ShapeNet by a large margin, particularly on CD and FID, which indicates a better perceivability. Further, we experiment on PointNet++ as well and demonstrate the results in Table \ref{Tab:PN++}. For PointNet++, we similarly select those layers with a large difference in activations when predicting real objects and raw AM instances, and heuristically set weights to minimize the discrepancies. The test results are consistent with PointNet in that non-generative AMs except Flow AM fail to generate global explanations with realistic object outlines.

\begin{table*}[]
\begin{tabular}{cccccc|ccc}
\hline
       & \multicolumn{5}{c|}{Non-generative model-based}                                                                           & \multicolumn{3}{c}{Generative model-based}                         \\ \hline
       & Vanilla                       & L2-norm              & Gaussian blur        & Total Variance       & Flow (ours)              & AE \cite{tan2023visualizing}                   & AED \cite{tan2023visualizing}                  & NAED \cite{tan2023visualizing}                 \\ \hline
Logit$\uparrow$  & 16.6                          & 6.3                  & 6.2                  & 6.0                  & \textbf{16.7}        & 10.5                 & 7.0                  & 8.1                  \\
Sftmax$\uparrow$ & $\mathbf{-4.5\times 10^{-4}}$ & $-4.1\times 10^{-3}$ & $-3.7\times 10^{-3}$ & $-9.8\times 10^{-3}$ & $-2.5\times 10^{-3}$ & $-9.1\times 10^{-2}$ & $-7.6\times 10^{-1}$ & $-6.0\times 10^{-1}$ \\
CD$\downarrow$     & 0.324                         & 0.139                & 0.148                & 0.376                & 0.081                & \textbf{0.044}       & 0.086                & 0.074                \\
FID$\downarrow$   & 0.176                         & 0.256                & 0.420                & 0.092                & 0.077                & 0.016                & 0.018                & \textbf{0.014}       \\ \hline
\end{tabular}
\caption{Quantitative evaluation of existing point cloud AM methods. The metrics from top to bottom are the magnitude of the neurons for corresponding classes in the Logits and SoftMax layers, the Chamfer distance, and the Fréchet inception distance, respectively.}\label{tab:quantitative_eva}
\end{table*}

\begin{table}[]
\centering
\begin{tabular}{cccccc}
\hline
                 & Vanilla              & L2              & GB        & TV                & Flow                    \\ \hline
Logit$\uparrow$  & 9.5                  & 14.6                 & 9.4                  & 7.8                           & \textbf{15.2} \\
Sftmax$(\times 10^{-1})\uparrow$ & $-13$ & $-29$ & $-25$ & $\mathbf{-3.8}$ & $-4.3$           \\
CD$\downarrow$   & 0.304                & 0.544                & 0.507                & 0.245                         & \textbf{0.067}                 \\
FID$\downarrow$  & 2.714                & 11.022               & 3.450                & 0.232                         & \textbf{0.033}                 \\ \hline
\end{tabular}
\caption{Quantitative evaluations for Non-generative model-based AM explanations on ShapeNet. AM regularization methods from left to right are: vanilla, L2-norm, Gaussian blur, total variation and our Flow AM.}\label{Tab:quantitative_shapenet}
\end{table}

\subsection{Threat from generative models: Sanity check}
\begin{figure}
    \begin{centering}
    \includegraphics[width=0.475\textwidth]{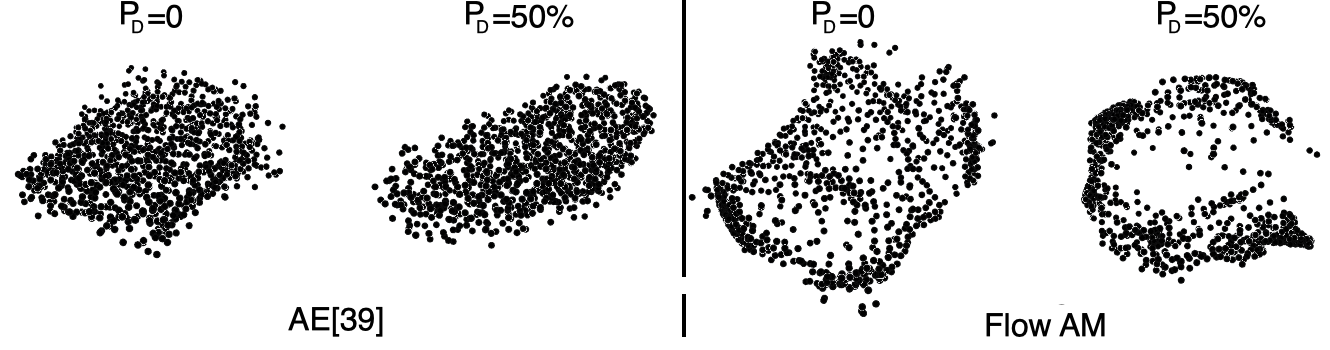}
    \caption{Sanity checks for generative and non-generative model based AM. $P_D$ represents the dropout probability, which denotes the original model when equals to zero. The two figures on the left and right are the explanations generated by AE and Flow AM, respectively.}
    \label{Fig:Sanity}
    \end{centering}
\end{figure}

Despite the higher quality of explanations generated by AM based on generative models, a potential concern is the fidelity to the model to be explained. Generative models are inherently strong reconstructionists, which interferes with the information source of the explanations and reduces their persuasiveness. We perform a sanity check to compare the fidelity of AM methods based on generative and non-generative models. We employ AE, i.e., the Autoencoder-based method proposed in \cite{tan2023visualizing}, and our Flow AM, respectively, to generate global explanations for the category ``bed" of PointNet trained on ModelNet40. Subsequently, we set a $50\%$ dropout probability for each parameter of the model. The dropout model almost loses its ability to classify due to sparsity and the absence of critical parameters. We again explain the dropout model with AE and Flow AM, respectively, and compare whether the explanations are corrupted after dropout. As shown in Fig. \ref{Fig:Sanity}, generative model-based AM remains capable of generating well-outlined global explanations when suffering dropouts and thus fails the sanity check. In contrast, Flow AM, where there is no external interference and all information is derived from the model, suffers from a severe deformation in the explanation after dropout and is no longer able to generalize the category ``bed". As a result, generative model-based AM may import extensive information from generative models, diminishing the fidelity of the model to be explained. Another observation is that the gradients of the generative model often conflict with the model to be explained, resulting in significantly lower activation magnitudes for the generative model-based explanations than for the Vanilla and flow AMs in Table \ref{tab:quantitative_eva}. More detailed analysis is presented in Section \ref{sec:more_sanity}.

\subsection{Ablation Study}
\begin{figure}
    \begin{centering}
    \includegraphics[width=0.475\textwidth]{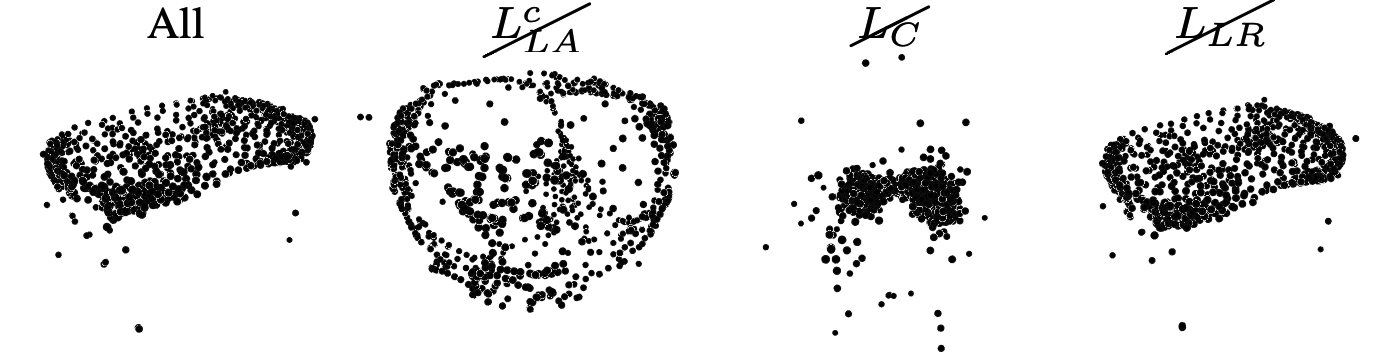}
    \caption{Visualization of ablation test. From left to right are Flow AM with all components included, with latent approximation removed, with coherence removed, and with legal constraints removed, respectively.}
    \label{Fig:Ablation}
    \end{centering}
\end{figure}

To disentangle and validate the effectiveness of individual components, we perform ablation tests on Flow AM-. The main refinements of Flow AM are the latent approximation, the point continuity and the legal restriction. In the ablation test, we remove one of the above terms individually, and evaluate the generated explanations. As shown in Figure \ref{Fig:Ablation}, eliminating latent approximation or point continuity drastically ruins the readability of the explanation from the perspective of humans. After ablating the latent approximation, the explanation is rendered spherical under the restraints of the point continuity and the outline of the object is lost. The removal of point continuity results in the majority of points shrinking and therefore losing semantics. This conclusion is also evidenced by the quantitative evaluation results in Table \ref{Tab:Ablation}. Additionally, though the legal restriction has insignificant impact on the quality of the explanations, its removal results in only $5\%$ of the explanations remain within the legal intervals of the dataset.

\begin{table}[]
\centering
\resizebox{0.475\textwidth}{10mm}{
\begin{tabular}{ccccc}
\hline
                    & All                  & $\cancel{L^c_{LA}}$  & $\cancel{L_{C}}$ & $\cancel{L_{LR}}$    \\ \hline
Logits$\uparrow$     & \textbf{16.7}                 & 14.9                 & -11.2            & 17.8                 \\
Sftmax$\uparrow$     & $\mathbf{-2.5\times 10^{-3}}$ & $-1.4\times 10^{-2}$ & -11.1            & $\mathbf{-2.5\times 10^{-3}}$ \\
CD$\downarrow$       & \textbf{0.081}                & 0.209                & /                & 0.087                \\
FID$\downarrow$      & \textbf{0.077}                & 0.114                & /                & 0.110                \\
Legality$\%\uparrow$ & $92.5\%$             & $\mathbf{100\%}$              & $\mathbf{100\%}$          & $5\%$                \\ \hline
\end{tabular}}
\caption{Quantitative evaluation of ablation test. From left to right are Flow AM with all components included, with latent approximation removed, with coherence removed, and with legal constraints removed, respectively. When $L_{C}$ is eliminated, we do not record the distance to real objects of the same class since all the explanations are misclassified by the model.}\label{Tab:Ablation}
\end{table}

\begin{table}[]
\centering
\resizebox{0.475\textwidth}{8mm}{
\begin{tabular}{cccccc}
\hline
                 & Vanilla              & L2              & GB        & TV       & Flow          \\ \hline
Logit$\uparrow$  & 6.9                  & 1.9                  & 2.0                  & 5.1                  & \textbf{15.4}        \\
Sftmax$\uparrow$ & $-4.0\times 10^{-4}$ & $-0.24$ & $-0.31$ & $-0.01$& $\mathbf{-5.4\times 10^{-7}}$ \\
CD$\downarrow$   & 0.116                & 0.272                & 0.225                & 0.066                & \textbf{0.049}       \\
FID$\downarrow$  & \textbf{0.013}                & 0.052                & 0.047                & 0.020       & 0.020       \\ \hline
\end{tabular}}
\caption{Quantitative evaluation of the global explanations of PointNet++ trained on ModelNet40.}\label{Tab:PN++}
\end{table}
\section{Limitations}
We show that Flow AM reconstructs well-outlined global explanations with information from the classifier without incorporating any generative model. However, there are still several limitations of Flow AM that need to be recognized.

\textbf{Diversity.} Diversity is one of the important elements of explanations that assist humans in analyzing models from various perspectives. We attempt to incorporate Gaussian noises in our experiments, but the noises significantly degrade the quality of the explanations, which renders it difficult to balance diversity and perceivability simultaneously. We demonstrate in Section \ref{sec:diversity} the incorporation of noises in different positions and the corresponding qualitative and quantitative results. Achieving diversity in global explanations without generative modeling might be a potential future work.

\textbf{Optimal parameter setting.} Flow AM exploits the differences of neuron activations in the intermediate layers for inputs with various outlines to force the generated explanations to approximate real objects, thus, accurately identifying and appropriately weighting those layers are crucial. We show the impact of fine-tuning $\beta$ on the quality of the generated explanations as an example in Sec. \ref{sec:hyperparameter}. However, we have not found a theoretically or empirically optimal weighting approach. All weights are set heuristically, which would be a significant increase in workload in the face of more complex models. A possible solution is to gain further understanding of how the intermediate layers learn contours, and the mechanisms by which they interact to adaptively set the weights.

Moreover, we observe that Flow AM only reveals promising results on point cloud models, while its effectiveness is limited for image models (See Sec. \ref{sec:compare_vs_image} for analysis). We consider this is because multi-size convolutional kernels share weights when learning local features and are struggling to be aligned.

\section{Conclusion}
This work aims to compensate for the lack of existing researches on the reliability of point cloud models by proposing a novel activation maximization method without incorporating generative models. Our method significantly improves the representativeness and perceptibility in comparison to other methods based on traditional regularization, which contributes to human understanding of black-box models. In future work, we attempt to promote the diversity of explanations as well as enable adaptive weight settings for more structurally complex models.
\clearpage
{
    \small
    \bibliographystyle{abbrv}
    \bibliography{main}
}

\clearpage
\beginsupplement
\section{Supplementary Material}

\subsection{Formulations of existing non-generative AM for images} \label{sec:tradictionAMformular}
In this section we present the traditional AM regularization methods that are applicable to images.
\textbf{L2-norm.} L2-norm is proposed to target extreme pixels that frequently occur in image AM. By weighting the gradients, L2-norm can effectively penalize the magnitude of extreme pixels in an image. L2-norm is formulated as

\begin{equation}
    x_{t+1}= x_t\cdot (1-\theta_{l2}) (x_t + \Delta x_t)
\end{equation}
where $x_t$ is the AM process at the $t^{th}$ iteration, $\theta_{l2}$ is an adjustable weight parameter and $\Delta x_t$ is the gradient of $x_t$.
\textbf{Gaussian blur.} Gaussian blurring incorporates a Gaussian kernel that filters each channel of the image to regularize extreme pixels, which is formulated as

\begin{equation}
    x_{t+1}= \frac{1}{\sqrt{2\pi \sigma ^2}}e^{-\frac{x^2}{2\sigma^2}}(x_t + \Delta x_t)
\end{equation}
where $\sigma$ is the standard diviation.

\textbf{Total Variation.} Total variance is a method of penalizing extreme points by regularizing the difference between neighboring pixels, which is formulated as

\begin{equation}
    x_{t+1} = (x_t + \Delta x_t)+\frac{\partial \underset{P}{\mathrm{sup}}\, \sum_{i=0}^{n_p-1}\left | x_t^{i+1} - x_t^{i} \right |}{\partial x_t}
\end{equation}
where $x_t^i$ is the $i^th$ pixel in $x_t$.
All of the above regularization methods aim to penalize extreme pixels, which is the most common issue in image AM. However, they do not apply to point cloud AM due to the disordered nature of the point cloud and the particular structure of the models.

\subsection{More visualizations}
\begin{figure*}
    \begin{centering}
    \includegraphics[width=1.0\textwidth]{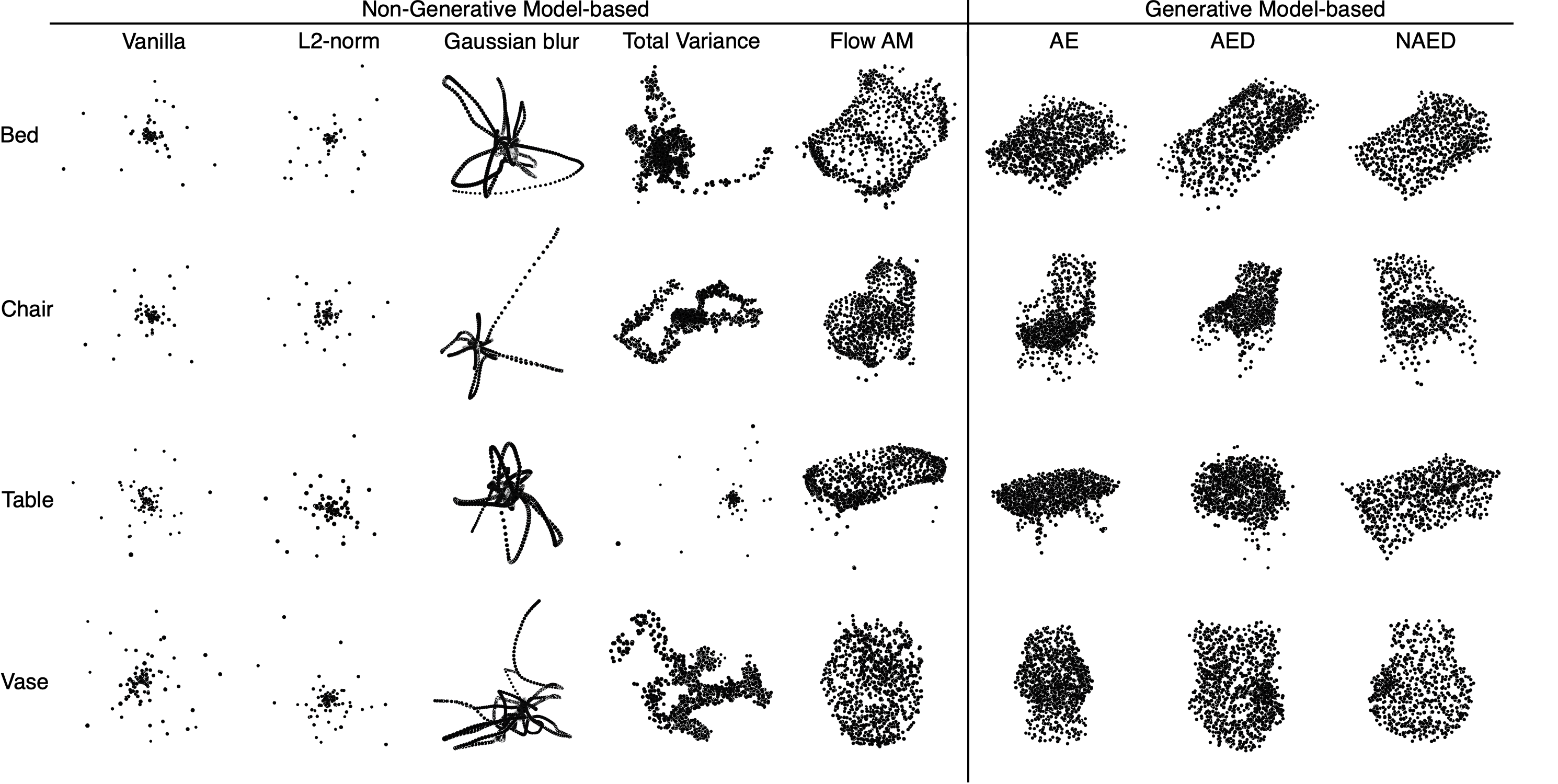}
    \caption{More visualizations. We show four additional common categories, from top to bottom: bed, chair, table and vase.}
    \label{Fig:more_visu_compare}
    \end{centering}
\end{figure*}

We show more visualizations of Flow AM and comparisons with other AM methods in Figure \ref{Fig:more_visu_compare}. It can be observed that Flow AM performs consistently on different classes, qualitatively approaching that of the generative model-based methods.

\subsection{More results of salinity checks} \label{sec:more_sanity}
\begin{figure*}
    \begin{centering}
    \includegraphics[width=1.0\textwidth]{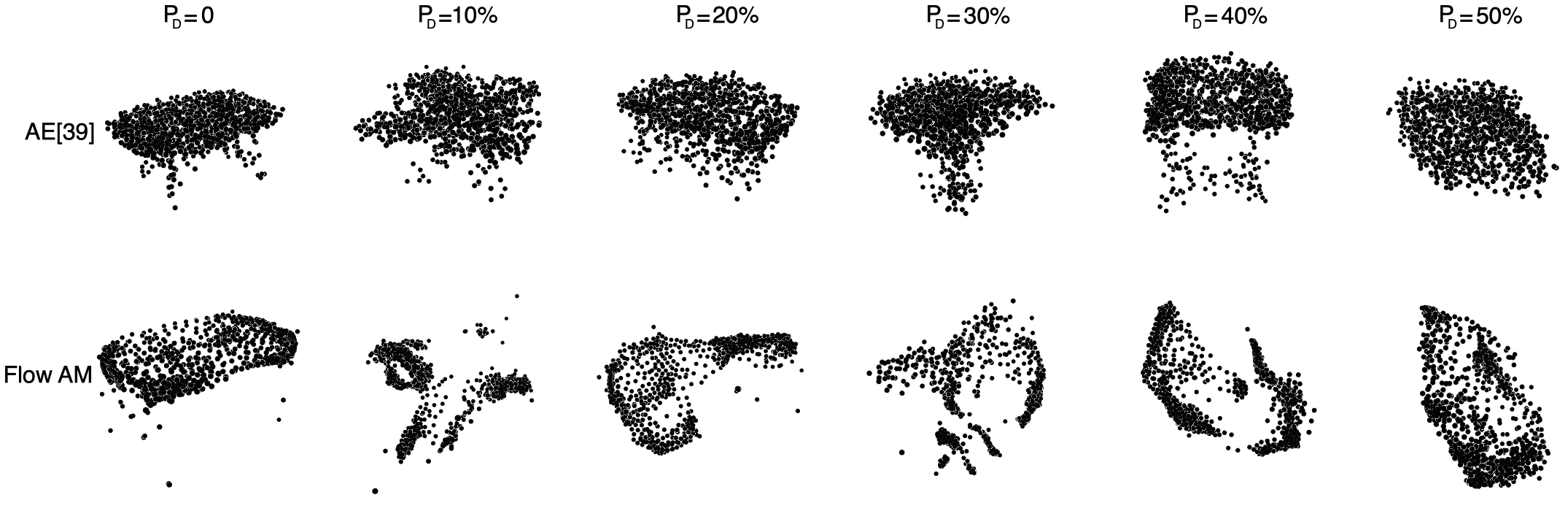}
    \caption{Complete salinity test results for the selected category ``Table". Above and below are the global explanations generated by AE from \cite{tan2023visualizing} and our proposed Flow AM, respectively. We show the extent to which the explanation collapses when the dropout probability is set from $0$ to $50\%$, respectively.}
    \label{Fig:more_sanity}
    \end{centering}
\end{figure*}

\begin{figure}
    \begin{centering}
    \includegraphics[width=0.475\textwidth]{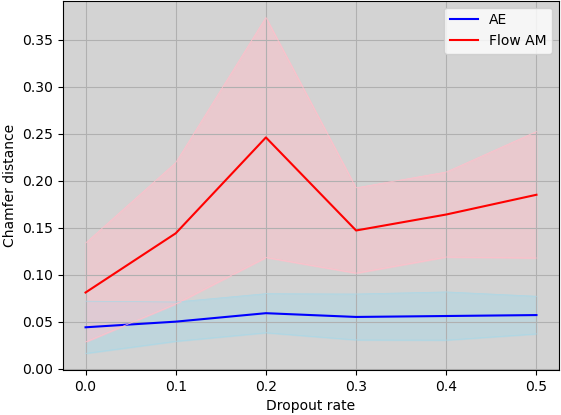}
    \caption{Mean and variance of Chamfer distance for explanations at different dropout rates.}
    \label{Fig:santy_stat}
    \end{centering}
\end{figure}

In this section we demonstrate the results of the salinity check in detail. Fig. \ref{Fig:more_sanity} illustrates the visualization of global explanations generated by AE \cite{tan2023visualizing} and Flow AM for the category "table" when the dropout rate is increased from $0$ to $50\%$. For AE, when the dropout rate rises to $40\%$, the outline of the table is still clearly recognizable, and even when the dropout rate is as high as $50\%$, the tabletop is still visible. In the case of Flow AM, on the other hand, whenever a small number of parameters are dropout, the profile of the explanation collapses completely. We again validate all classes for the entire dataset and present the quantitative results in Table \ref{tab:sanitycheck2} and Fig.\ref{Fig:santy_stat} . Note that for fairness, the latent vector similarity FID is ignored because the latent distance measured by the point cloud FID is exactly in the optimization objective of Flow AM. The quantitative results indicate that the generative model-based approach AE does not exhibit a significant quality collapse when as the dropout rate increases. In contrast, the point by point similarity of Flow AM grows dramatically at a dropout rate of $10\%$ and is extremely unstable (red curve in Fig. \ref{Fig:santy_stat}). The results suggest that the contour information of the generative model-based AM method mainly originates from the internal of the generative model rather than the model to be explained. Although generative models noticeably enhance the quality of explanations, they should be applied with caution for reasons of fidelity.

\begin{table*}[]
\centering
\begin{tabular}{ccccccc}
\hline
                                              & $P_D=0$ & $P_D=0.1$ & $P_D=0.2$ & $P_D=0.3$ & $P_D=0.4$ & $P_D=0.5$ \\ \hline
AE \cite{tan2023visualizing} & 0.044   & 0.050     & 0.059     & 0.055     & 0.056     & 0.057     \\
Flow AM (ours)                                & 0.081   & 0.144     & 0.246     & 0.147     & 0.164     & 0.185     \\ \hline
\end{tabular}
\caption{Quantitative evaluation of salinity checks. Quantitative evaluation results of the explanations generated by the model at different dropout rates. For fairness, we only choose Chamfer distance as the evaluation metric since Flow AM is intended to approximate the latent distance, and thus FID does not reflect the true perceptual differences.}\label{tab:sanitycheck2}
\end{table*}

\subsection{Adding noises for diversity} \label{sec:diversity}
We consider the following three types of noise adding:

\textbf{Initialization.} The most intuitive solution is to add noise to the initializations. Recall that we start with an initialization of the mean values of the same categories in the dataset, and in order to reach the final diversity of explanations, adding an appropriate amount of noise intuitively enables the optimization to start from a different beginning without compromising too much on performance.

\textbf{Latent activation.} We have found that the outline of the explanations is determined by the activation of the intermediate layers. Thus, incorporating a moderate amount of noise on the latent activations prevents the explanations from being consistently directed toward the same endpoint. However, due to the limitations of setting weights heuristically, identifying suitable noise weights is equally challenging.

\textbf{Iteration process.} Another intuitive way to avoid homogenization of the explanations is to add quantified noise at each step of the optimization process. The risk of adding noise iteratively is to render the optimization process difficult to converge and degrade the quality of the explanations.

\begin{figure*}
    \begin{centering}
    \includegraphics[width=1.0\textwidth]{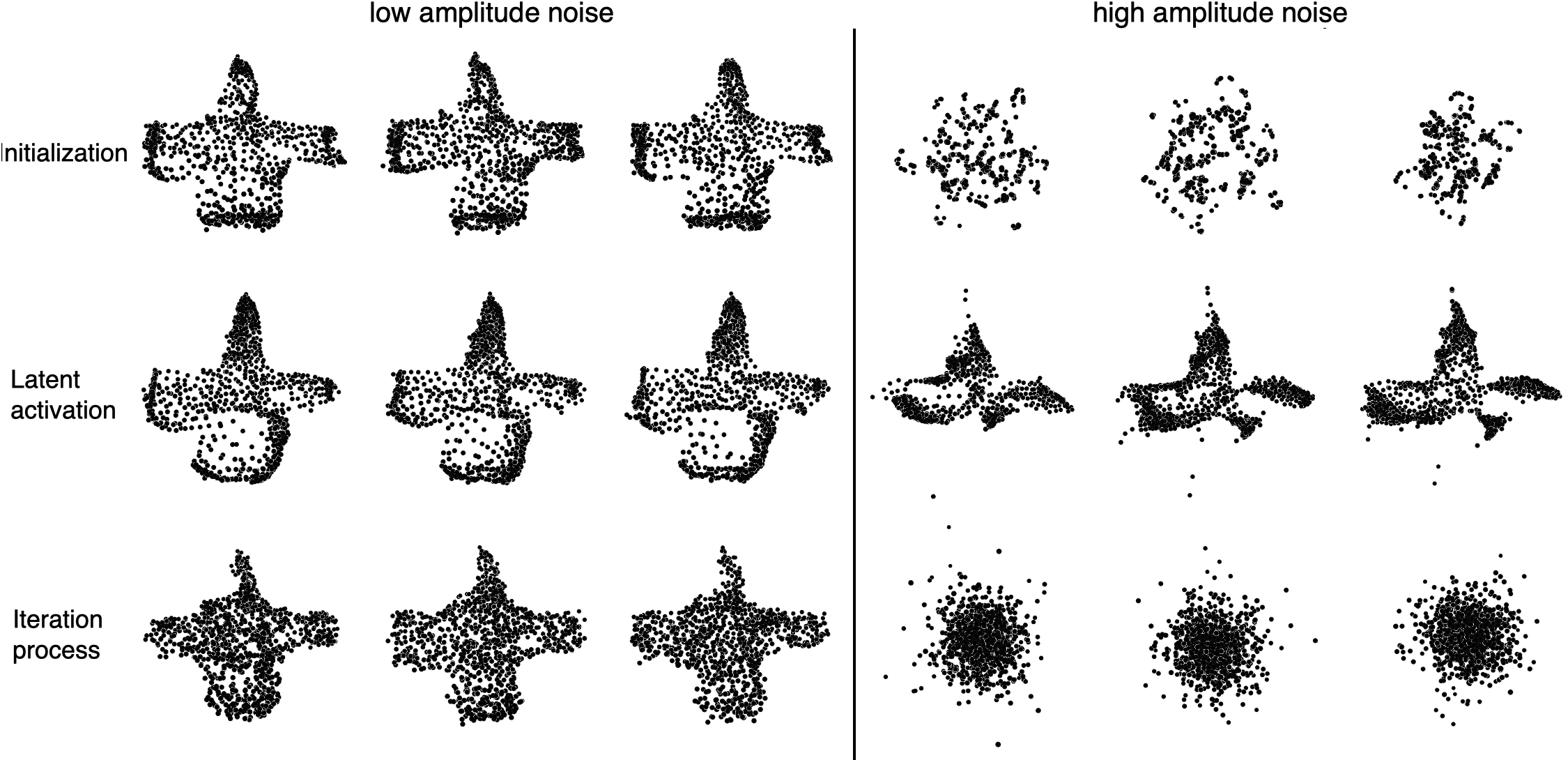}
    \caption{Three different approaches to noise addition dedicated to explaining diversity. The left and right sides show the addition of low-amplitude and high-amplitude noise to the corresponding mechanisms, respectively.}
    \label{Fig:diversity}
    \end{centering}
\end{figure*}

Fig. \ref{Fig:diversity} illustrates the global explanations generated by the three aforementioned noise additions. We add two types of noise to the above structure, i.e., low-amplitude and high-amplitude, which are Gaussian noise with small and large variance (relative to the tensor being noised), respectively. However, when the noise amplitude is relatively small, the above methods fail to produce diverse explanations. When large amplitude noise is added to the initial input or the iterative process, although the explanations show a certain degree of diversity, the overall contour of the object is completely corrupted.

\subsection{Impact of fine-tuning hyperparameters on explanations} \label{sec:hyperparameter}

\begin{figure}
    \begin{centering}
    \includegraphics[width=0.475\textwidth]{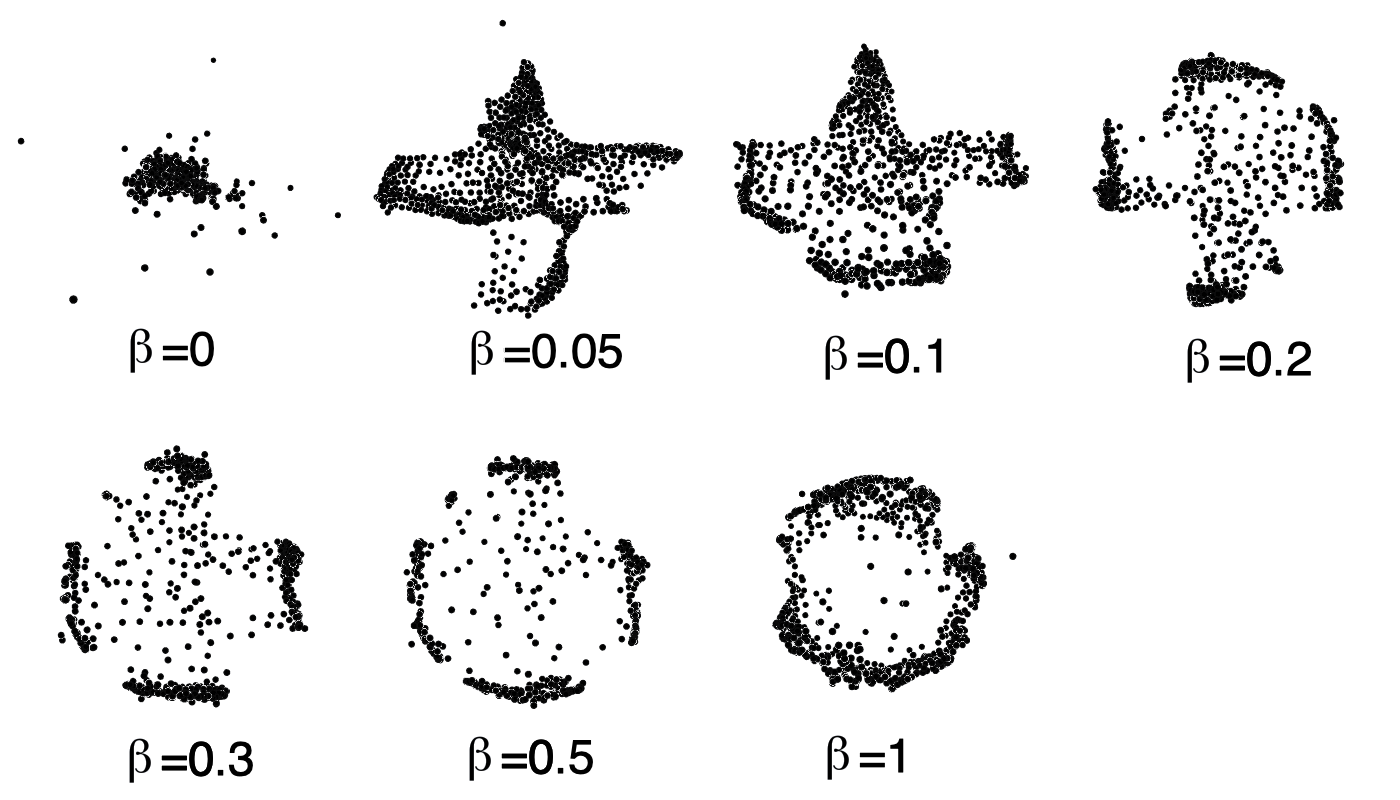}
    \caption{The impact of $\beta$ on the quality of explanations. We set $\beta$ as $0$, $0.05$, $0.1$, $0.2$, $0.3$, $0.5$ and $1$, respectively}
    \label{Fig:hyperparameter}
    \end{centering}
\end{figure}

Hyperparameters affect the quality of generated explanations to a large extent. We take $beta$ as an example, which is a hyperparameter for balancing latent alignment and point continuity. We generate explanations for ``airplane" categories on PointNet trained on ModelNetl40, but observe the impact on the quality of the explanations by adjusting the value of $beta$, and present the results in Fig. \ref{Fig:hyperparameter}. When $\beta$ is excessively small, although the critical points are sufficient to summarize the outline of the object, the quality of the explanation is too low due to the fact that the rest of the points are not expanded enough to be recognized by humans. For the selected model, $\beta=0.1$ is a relatively preferable tradeoff. As $\beta$ continues to increase, the regularization places more emphasis on point continuity, which in the limit of the gradient causes almost all points to cluster next to the critical points, leading to the formation of a vacuum in the middle of the object. Note that $\beta$ and other hyperparameters are not generalizable, i.e., values that apply to PointNet may not be applicable to other models. Thus for each model, the hyperparameters need to be tuned through extensive experiments to generate high-quality global explanations.

\subsection{Specificity of point cloud models} \label{sec:compare_vs_image}

\begin{figure}
    \begin{centering}
    \includegraphics[width=0.475\textwidth]{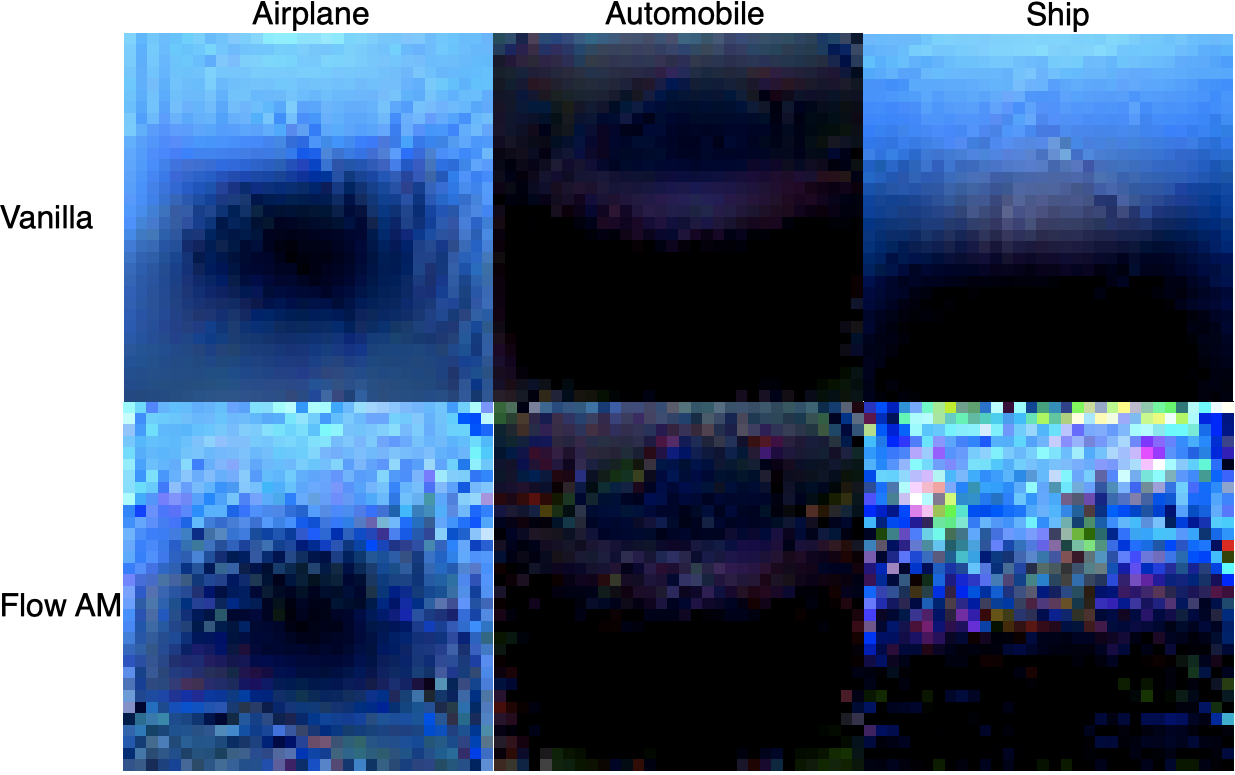}
    \caption{Comparison of global explanations generated by Vanilla and Flow AM on ResNet18. The top and bottom rows are Vanilla and Flow AM, respectively, and the categories from left to right are ``airplane", ``automobile", and ``boats". }
    \label{Fig:2D_flow}
    \end{centering}
\end{figure}

\begin{figure*}
    \begin{centering}
    \includegraphics[width=1.0\textwidth]{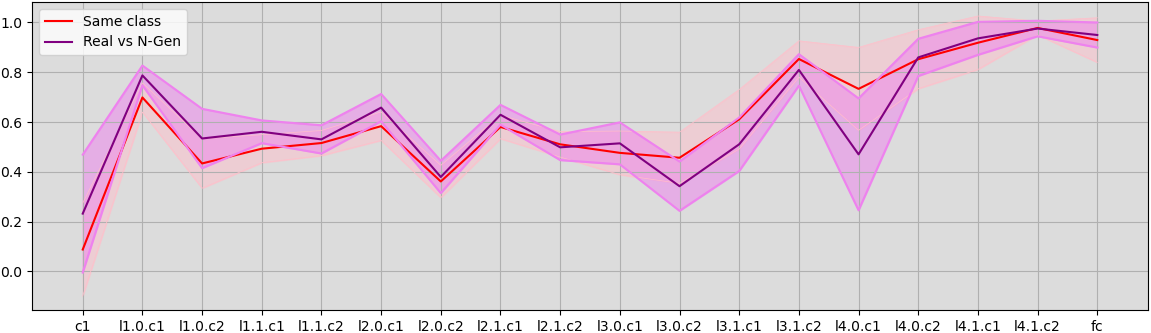}
    \caption{Comparison of latent distances between global explanations generated by vanilla AM and real objects for image model. The model is ResNet18 trained on CIFAR10 dataset. $l$ and $c$ in the x-axis represent layers and convolutional layers, respectively.}
    \label{Fig:2D_activation}
    \end{centering}
\end{figure*}

We attempt to verify whether the proposed method is also applicable to image models. We train a ResNet18 on CIFAR10, which achieves $93.2\%$ accuracy on the test set. We follow the process in Fig. \ref{Fig:Flow_overview} and generate global explanations with Flow AM. Nevertheless, as shown in Fig. \ref{Fig:2D_flow}, we observe that Flow AM does not enable the image model to generate higher quality global explanations. We attribute this to the structural differences between images and point cloud models. We show in Fig. \ref{Fig:2D_activation} the latent distance difference between vanilla explanations and real objects in the image model. Remarkable discrepancies in latent distances exist in the convolutional layers (e.g. $l3.0.c2$ and $l4.0.c1$) rather than in the fully-connected layers as in the point cloud models. This is because the multi-size convolutional kernel plays the role of learning the correlations of the pixels in the field of view of the image through which the predictions are made. However, aligning the latent distances in convolutional kernels is problematic, as is aligning point-wise convolutions in point cloud models, which share the same weights but are required to learn different local features. Therefore, the proposed method is not applicable to image models where multi-size convolutional kernels serve as the main framework.

\end{document}